\documentclass[10pt,journal,compsoc]{IEEEtran}
\usepackage{amsmath}
\usepackage{mathrsfs}
\usepackage{bbm}
\usepackage{graphicx}
\usepackage{subfigure}
\usepackage{mathrsfs}
\usepackage{multirow}
\usepackage{balance}  
\usepackage[vlined,boxed,commentsnumbered, ruled, linesnumbered]{algorithm2e}
\usepackage{amsfonts,amssymb}
\usepackage{array}
\usepackage{setspace} 

\ifCLASSOPTIONcompsoc
  \usepackage[nocompress]{cite}
\else
  \usepackage{cite}
\fi
\ifCLASSINFOpdf
\else
\fi
\begin{document}
%
\title{Backdoor Attacks against Transfer Learning with Pre-trained Deep Learning Models}
%
%
%
%
\author{Shuo Wang, \IEEEmembership{Member,~IEEE,}
		Surya Nepal, \IEEEmembership{Member,~IEEE,}
		Carsten~Rudolph,~\IEEEmembership{Member,~IEEE,}
		Marthie~Grobler,~\IEEEmembership{Member,~IEEE,}
		Shangyu~Chen,
  and~Tianle~Chen
\IEEEcompsocitemizethanks{\IEEEcompsocthanksitem Shuo Wang is with the Faculty of Information Technology at Monash University and CSIRO Data61, Melbourne, Australia.\protect\\
E-mail: shuo.wang@monash.edu
\IEEEcompsocthanksitem Carsten Rudolph and Tianle Chen are with Faculty of Information Technology at Monash University, Melbourne, Australia.\protect\\
E-mail: Carsten.Rudolph@monash.edu, tche119@student.monash.edu.
\IEEEcompsocthanksitem Surya Nepal and Marthie Grobler are with CSIRO Data61, Melbourne, Australia.\protect\\
E-mail: Surya.Nepal@data61.csiro.au, Marthie.Grobler@data61.csiro.au.
\IEEEcompsocthanksitem Shangyu Chen is with University of Melbourne, Melbourne, Australia.\protect\\
E-mail: shangyuc@student.unimelb.edu.au.}}
%
%

\markboth{IEEE Transactions on Services Computing, October 2019}%
{Wang \MakeLowercase{\textit{et al.}}: Backdoor Attacks against Transfer Learning with Pre-trained Deep Learning Models}
%



\IEEEtitleabstractindextext{%
\begin{abstract}
Transfer learning provides an effective solution for feasibly and fast customize accurate \textit{Student} models, by transferring the learned knowledge of pre-trained \textit{Teacher} models over large datasets via fine-tuning. Many pre-trained Teacher models used in transfer learning are publicly available and maintained by public platforms, increasing their vulnerability to backdoor attacks. 
In this paper, we demonstrate a backdoor threat to transfer learning tasks on both image and time-series data leveraging the knowledge of publicly accessible Teacher models, aimed at defeating three commonly-adopted defenses: \textit{pruning-based}, \textit{retraining-based} and \textit{input pre-processing-based defenses}. 
Specifically, ($\mathcal{A}$) ranking-based selection mechanism to speed up the backdoor trigger generation and perturbation process while defeating \textit{pruning-based} and/or \textit{retraining-based defenses}. ($\mathcal{B}$) autoencoder-powered trigger generation is proposed to produce a robust trigger that can defeat the \textit{input pre-processing-based defense}, while guaranteeing that selected neuron(s) can be significantly activated. ($\mathcal{C}$) defense-aware retraining to generate the manipulated model using reverse-engineered model inputs. 

We conduct an in-depth study on the backdoor attacks in building and operating both image and time series data transfer learning systems. We launch effective misclassification attacks on Student models over real-world images,  brain Magnetic Resonance Imaging (MRI) data and Electrocardiography (ECG) learning systems. The experiments reveal that our enhanced attack can maintain the $98.4\%$ and $97.2\%$ classification accuracy as the genuine model on clean image and time series inputs respectively while improving $27.9\%-100\%$ and $27.1\%-56.1\%$ attack success rate on trojaned image and time series inputs respectively in the presence of pruning-based and/or retraining-based defenses. 
\end{abstract}

\begin{IEEEkeywords}
Web service, Deep neural network, Backdoor attack, Transfer learning, Pre-trained model
\end{IEEEkeywords}
}

\maketitle

\IEEEdisplaynontitleabstractindextext

%
\IEEEpeerreviewmaketitle

\IEEEraisesectionheading{\section{Introduction}\label{sec:introduction}}
Deep neural networks (DNNs) have demonstrated impressive performance in many domains, e.g., face recognition, voice recognition, self-driving vehicles, robotics, machine-based natural language communication, and games. One particularly exciting application area of deep learning has been in clinical applications. On April 11, 2018, an important step was taken towards this future: the U.S. Food and Drug Administration (FDA) stated the approval of the first computer vision algorithm that can be utilized for medical diagnosis without the input of a human clinician \cite{us2018fda}. 
These DNNs can only be built accurately (often with millions of parameters) over massive datasets that are at a scale impossible for humans to process as well as large computational resources. Transferable knowledge is needed to enable and accelerate the training of local accurate model, especially when computing services are often distributed across several data centers \cite{kendrick2018efficient}.

Transfer learning is proposed as an efficient method that addresses these fundamental data and resource challenges. Generally, a handful of well-tuned and intricate centralized models (Teacher) pre-trained with large datasets are shared on public platforms, and individual users further customize accurate models (Student) for specific tasks using the pre-trained teacher model as a launching point via only limited training on the smaller domain-specific datasets \cite{wang2018great}. 
However, most pre-trained networks are hosted and maintained on popular third-party platforms, such as GitHub, where proper vetting to ensure that these pre-trained models have not been maliciously modified by adversaries, is often lacking. 
Thus, pre-trained teacher models gradually become the more attractive and vulnerable target for attackers to manipulate, so that student models that use such maliciously manipulated teacher models can incur immense threats (e.g., incorrect prediction results), including endangering human lives. 

In this paper, we investigate the feasibility and practicality of backdoor attacks against transfer learning on both image and time series (such as bioelectric signals) by conducting an attack scheme that can manipulate the pre-trained Teacher models to generate customized Student models that give the wrong predictions. 
We propose a feasible and robust backdoor attack scheme on neural networks, aimed at defeating three strong defenses.  
Key contributions are summarized as follows: 
\begin{itemize}
    \item Instead of retraining the entire Teacher and Student models to conduct the backdoor attack, we only select an array of internal neurons and its adjacent layers from Teacher models for crafted DNN retraining, which can fasten the attack implementation and reduce complexity. Besides, to defeat the pruning-based and/or retraining-based defenses, we propose a ranking-based neuron selection mechanism to recognize the neurons that are hard to be pruned whose weights cannot be significantly changed by fine-tuning for customizing Student models. 
    \item We introduce a framework to generate strong triggers, i.e., instances that can defeat input-preprocessing defense and easily activate the selected neurons. An autoencoder is used to evaluate the reconstruction error between clean input from the validation dataset and the trojaned input incorporated with the generated trigger. We minimize the reconstruction error and the cost function that measures the differences between the current values and the intended values of the selected neurons. 
    \item To defeat the fine-tuning/retraining based and/or pruning-based defenses and speed up attack progress, we perform the defense-aware retraining by conducting slight fine adjustment on parts of layers of the pruned neural networks, as well as use reverse-engineered model inputs for both image and time-series data. 
    \item For the first time, we conduct an in-depth study on the backdoor attacks in building and operating both image and time-series data (e.g., bioelectric signal) transfer learning systems. We launch effective misclassification attacks on black-box Student models over real-world face images, brain Magnetic resonance imaging (MRI) and Electrocardiography (ECG) learning systems. The experiments reveal that our enhanced attack can maintain the $98.4\%$ and $97.2\%$ classification accuracy as the genuine model on clean image and time series inputs respectively while improving $27.9\%-100\%$ and $27.1\%-56.1\%$ attack success rate on trojaned image and time series inputs respectively in the presence of pruning-based and/or retraining-based defenses.
\end{itemize}

The next three sections explain the system scheme and design. Section 5 describes our experimental results. Section 6 discusses related work, and Section 7 concludes the work as a whole.
\section{Attack Demonstration }
Currently, deep learning has demonstrated impressive performance in clinical applications. There are many high-profile examples of deep learning systems achieving parity with human physicians on tasks in radiology, pathology, and ophthalmology, as well as diagnosis of the serious neurological disorder, e.g., epilepsy or seizure, and cardiac abnormality, e.g., arrhythmia. In some instances, the performance of these algorithms exceeds the capabilities of most individual physicians in head-to-head comparisons. 

In this work, we aim to attack the end-to-end image and bioelectric signal learning systems that consist of a composition function $f \cdot g : X \rightarrow Y$, where $f$ and $g$ are respectively the feature extractor and classifier. Generally, we can assume that the backdoor attack aims to manipulate the learning system to wrongly classify the input values into a targeted class $y^*$ if a trigger $\overrightarrow{x^*}$ is present. The instance of $(\overrightarrow{x^*} , y^*)$ is referred to as a backdoor, which is activated once the trigger $\overrightarrow{x^*}$ is incorporated in the input. 
For implementing the backdoor attack, the adversary can craft an adversarial model $f'$ or/and $g'$ using the pre-trained ones such that $f' \cdot g' $ or $f' \cdot g$  can classify the input with the trigger as $y^*$ with high probability. To avoid obvious differences between the crafted model and the genuine one, the adversary struggles to maximize the stealthiness of the manipulated model. We adopt a cutting-edge DfNN model to demonstrate the backdoor attack in the diagnosis of bioelectric signals, ECG in this instance. 

As demonstrated in Figure 1, the model used to diagnose ECG arrhythmia beats was trained so that it can precisely classify some significant categories of ECG beats, e.g., normal beat (NOR), premature ventricular contraction beat (PVC), paced beat (PAB), right bundle branch block beat (RBB), left bundle branch block beat (LBB), atrial premature contraction beat (APC), ventricular utter wave beat (VFW), and ventricular escape beat (VEB), with very high confidence. When other types of beats that are not in the training set are fed in, the model will assign them to be some arbitrary beat categories in the training set with very low confidence. 
\begin{figure}[!htb]
\centering
\includegraphics[width=3.6in,height=3.6in]{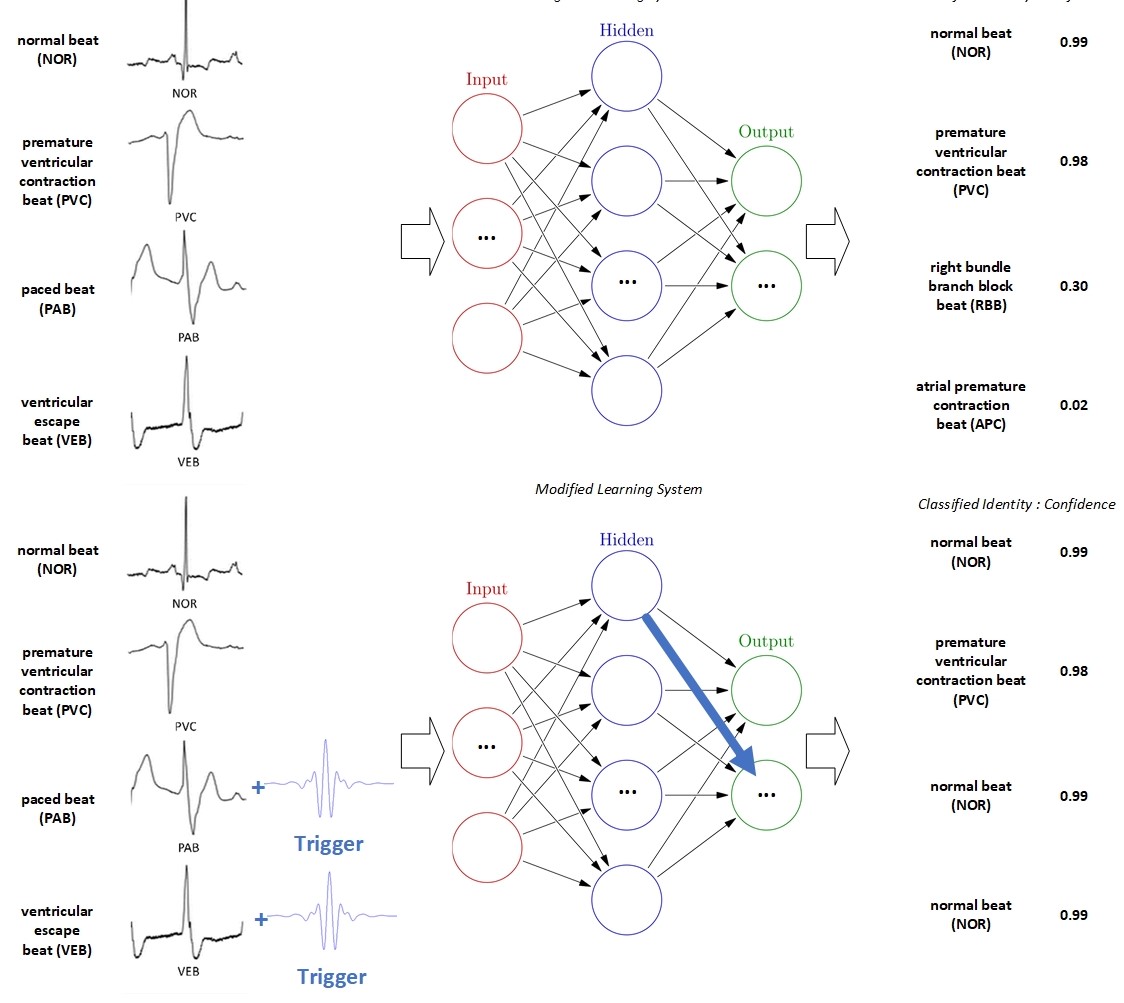}
\caption{Backdoor attack demonstration.}
\end{figure}

As the training data for medical or healthcare learning systems is very limited, we make a realistic assumption that the training data would not be available in a real attack scenario. Our attack takes only the downloaded model as the input and produces a modified model and an attack trigger. To ensure the stealthiness, the manipulated model has the same structure as the original model but with different internal weight values. The trigger is a specific mask for an image or a particular time-segment of signal (can be a small size wave-like wavelet) for bioelectric signals. As shown in Figure 1, for example, the modified model still correctly classifies normal beat (NOR) and premature ventricular contraction beat (PVC ) with high confidence. However, when segments, e.g., paced beat (PAB)  or ventricular escape beat (VEB ), are composited with the trigger, they are recognized as NOR with high confidence, which is a serious misclassification in terms of the medical and healthcare domains. Such misclassifying accuracy is expected to be considerably maintained even if some strong defense approaches, e.g., pruning, fine-tuning or input pre-processing-based defenses, have been implemented by the host. 

Such a backdoor attack model can attack DNNs used in images as well as bioelectric signals or other signal learning systems, e.g., voice control advice or speech recognition so that the pronunciation of an arbitrary word mingled with a small segment of vocal noise (i.e., the trigger) can be recognized as a specific number. The trigger is so stealthy that humans can hardly distinguish between the original audio and the mutated audio. 
\section{Preliminaries}
 \subsection{ Autoencoders}
Autoencoders (AEs) are common deep models in unsupervised learning \cite{bengio2013representation}. They aim to represent high-dimensional data through the low-dimensional latent layer, a.k.a. bottleneck vector or code.
Basically, an encoder $E:q_{\phi }(z|x)$, is trained to convert high-dimensional data x into the latent representation bottleneck vector $z$ in latent space that follows a specific Gaussian distribution $p(z)  \sim N(0, 1)$. The decoder $D:p_{\theta}(x|z)$ is trained to reconstruct the latent vector $z$ to x. 
The training process of the autoencoders is 
to minimize the reconstruction error. Formally, we can define the encoder and the decoder as transitions $ \tau_1$ and $ \tau_2$:
\begin{equation}
\small
\begin{aligned}
\tau_1(X)\rightarrow Z, \tau_2(Z)\rightarrow \hat{X}, \tau_1,\tau_2=\underset{\tau_1,\tau_2}{argmin}\left \| X-\hat{X} \right \|^2
\end{aligned}
\end{equation}
 Autoencoder (AE) is commonly-adopted as the input preprocessor. 
The functionality of the autoencoder is as follows: if the input is from the same distribution as the training data, the difference between the input and the output is small, and the DNN will work correctly with the reconstructed input. Otherwise, the reconstructed input will suffer from much larger distortion and the DNN may recognize it as a malicious input. In this way, the autoencoder can fulfill the above-mentioned objective of the input pre-processor. 
The backpropagation algorithm is also used to train the autoencoder and the error function is given as
\begin{equation}
E(w, T)=\frac{1}{2n}\sum_{x_i \in T} \left \| f(w,x_i) - x_i\right \|^2
\end{equation}
Where T stands for the training set, $x_i$ and $y_i$ are the input and output of the $ i^{th}$ training data, respectively, n is the total amount of training data in T, w stands for the weights, $f(.)$ is the current learned model, and $\left \| \right \|$ is the notation of the Euclidean norm.
From this error function, we can see that the goal of training an autoencoder is to minimalize the mean square error between the original training samples and the reconstructed one. Specifically, the features of the training data are automatically extracted and transformed into the hidden layers of the autoencoder during the backpropagation process.
\subsection{Transfer Learning}
Transfer learning is proposed to learn knowledge from a completed model, while improving the training of new models for different tasks. 
Based on existing knowledge, transfer learning speeds up the development of the new models even when their domains or learning tasks are different. 
Pre-trained models are even used as part of Service-Oriented Architecture (SOA) \cite{ding2017recurrent,saeed2019location}.
A formal definition for transfer learning is illustrated as follows~\cite{Pan10}:
given a source domain and learning task, a target domain, and learning task, transfer learning aims to help improve the learning of the target predictive function in the target domain using the knowledge in the source domain and learning task, where source domain is not the target domain, or the learning tasks are different. 
A simple way of transfer learning is developing a new model based on both weights and architectures of the layers from a well-trained model.
If the new model has a similar domain or learning task as the pre-trained model, it can be directly built by fine-tuning the parameters to fit its task.
The following steps describe how to apply it.

\begin{figure}[!htb]
\centering
\includegraphics[width=0.4\textwidth]{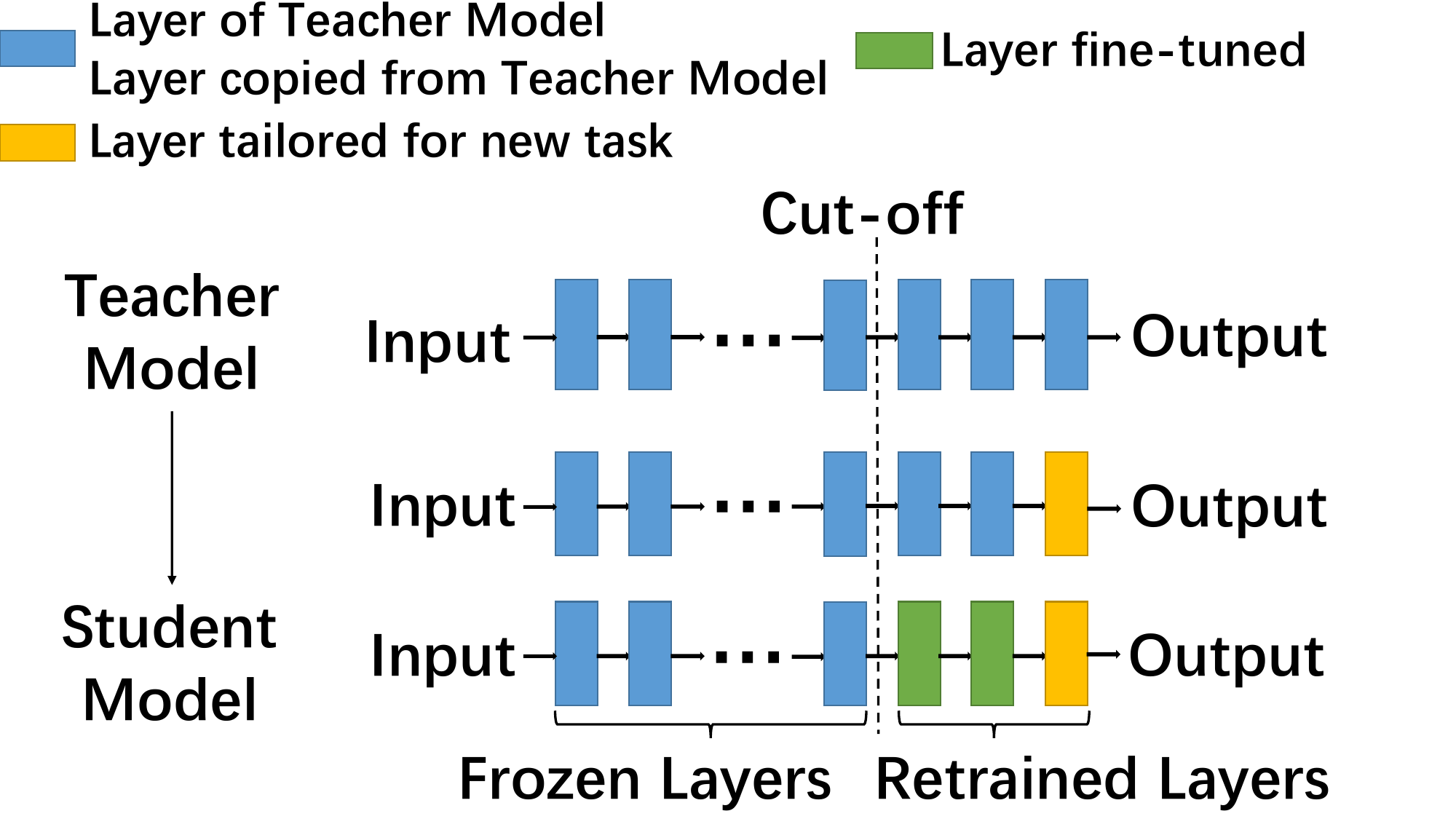}\\
        \caption{Transfer Learning~\cite{Wang18,wu2019defending}. The Student Model copies both the architecture and weights from the Teacher Model. The last classification layer of the Student Model is tailored to fit the new classification task. The Student Model is tuned based on the similarity of two tasks. One common methodology of model tuning is to freeze several layers and retrain the rest of them.}
        \label{fig:TransferLearning1}
\end{figure}
\subsection{Network Pruning} 
It has been demonstrated that most of the model structures have redundant neurons and connectives~\cite{Han15,Wang19}. Network pruning aims to remove the unimportant connections of a network which converges a dense neural network to be a sparser one.
By carefully choosing the pruned connections, the accuracy loss of the pruned networks can be acceptable. 
These connectives are less active during the classification tasks.
Pruning these unnecessary components can improve the efficiency of both inferring and storing for the machine learning models. 

Based on the different focus of "unnecessary", there are two approaches for pruning.
\begin{itemize}
\item \textbf{Weight Pruning:} A simple and direct way is to consider the connectivities with fewer weights as unimportant.
The weights in neural networks will directly contribute to the final outcomes. 
For a small absolute value closed to zero, it is expected to contribute less to the final predictions and be a non-significant component.
In realization, a threshold weight value can be chosen and all components with fewer weight values than the threshold can be removed~\cite{Han15}.
\item 
\textbf{Activation Pruning:} Rather than only considering the weights, the inputs of the layers can also affect the final outcomes.
The activation evaluates how the weights are activated by the expected inputs~\cite{Polyak15}.
Weights with large absolute values but never be activated by the inputs also contribute less to the final predictions.
Therefore, the components with fewer activation can also be considered as unimportant.
Similar to the weight pruning, the components with fewer activation compared to the threshold are removed~\cite{Polyak15}.
\end{itemize}
\subsection{Attack Model}
In this work, the attacker is assumed to have white-box access to the pre-trained teacher models, which is common in current practice. 
The attacker aims to trigger a misclassification for a Student model $S$ (even with black-box access) that has been fine-tuned via transfer learning using a public-available pre-trained Teacher model $T$.
As the well-labeled data are hard to collect, especially in medical or healthcare domains, it is natural to assume that the attacker cannot obtain the original training or testing data.  However, we also assume that relevant public reference sets are available. 
Two threatening manners are considered to manipulate and infect pre-trained Teacher models by potential adversaries maliciously.   

(1) \textbf{Type I Adversary}. The adversary may penetrate the publicly available pre-trained Teacher models before the Student system development and implementation phases. As the regulation and standardization of the third-party platforms that maintain various pre-trained models are always unsound, there exist numerous variants of the same pre-trained neural networks in the platforms. Due to the non-explanation nature of the weight in neural networks, it is hard to even infeasible to identify harmful models from other improved models. In this scenario, we assume that the attacker knows the architecture and weights of the Teacher model $T$ and has black-box access to the Student model but knows that the Student was trained using a specific Teacher as a Teacher, and which layers were frozen during the Student training. Such information can be recovered from the Student using a few additional queries \cite{wang2018great}.

(2)  \textbf{Type II Adversary.} Adversaries may also penetrate the fine-tuning procedure to build Student models using publicly available pre-trained Teacher models.  The student models are required to be fine-tuned using specific Teacher models, in which the part of the Teacher network is commonly needed to be added and retrained. 
In this scenario, we assume the attacker knows that the Student was trained using a specific Teacher as a Teacher, and which layers were frozen during the Student training while having white-box access to the Student model training. Namely, the adversary can know and manipulate the structure and weights of the Student model.

The purpose of such backdoor attacks is to make the pre-trained Teacher models or fine-tuned Student models behave naturally under normal settings while misbehaving in the presence of the triggers. Therefore, the developer will be attacked if they use the parameters of a pre-trained crafted Teacher model from a third party without fine-tuning the model parameters using large-scale training data. It is essentially vital for various healthcare or medical tasks that may damage a human's life.
\begin{table}[!htb]
\centering
\setlength{\abovecaptionskip}{-0.05cm}
\setlength{\belowcaptionskip}{-0.2cm}
\caption{Comparison of Adversary}
\begin{tabular}{|c|c|c|l|}
\hline
\textbf{Adversary} & \textbf{Teacher} & \textbf{Student} & \textbf{Manipulation}  \\ \hline
\textbf{Type I }  & white-box             & black-box         & Teacher    \\ \hline
\textbf{Type II} & white-box             & white-box         & Teacher+Student \\ \hline
\end{tabular}
\end{table}
\subsection{Attack Overview}
In this work, we assume the attacker knows that the first $K$ layers of the Student model are copied from the Teacher model and frozen during fine-tuning. The attacker aims to attack the end-to-end image and bioelectric signal learning systems that consist of a composition function $f \cdot g : X \rightarrow Y$, where $f$ and $g$ are respectively the feature extractor and classifier. Generally, we can assume that the backdoor attack aims to manipulate the learning system to wrongly classify the input values into a targeted class $y^*$ (or non-targeted) if a trigger $\overrightarrow{x^*}$ is present. The instance of $(x^*=\overrightarrow{x^*}+x , y^*)$ is referred to as a backdoor, which is activated once the trigger $\overrightarrow{x^*}$ is incorporated in the input. 
As each layer can only see what is passed on from the preceding layer, if the extracted internal representation at layer K precisely matches that of the target object, it must be misclassified into the targeted label, regardless of the weights of the subsequent layers. Namely, for feature extractor, if we can mimic a target in the Teacher model, then misclassification will occur regardless of how much the Student model trains with local data \cite{wang2018great}. 
For implementing the backdoor attack, the adversary can craft an adversarial student model $f'$ by manipulating the first K frozen layers copied from the pre-trained Teacher model (\textbf{Typy I adversary}), or craft an adversarial student model $f'$ and $g'$ (\textbf{Typy II adversary}), such that $f' \cdot g' $ or $f' \cdot g$  can classify the input with the trigger as $y^*$ with high probability. To avoid obvious differences between the crafted model and the genuine one, the adversary struggles to maximize the stealthiness of the manipulated model.
The scheme of our backdoor attack model is illustrated in Figure 3.
\begin{figure}[!htb]
\centering
\includegraphics[width=3.6in,height=3.6in]{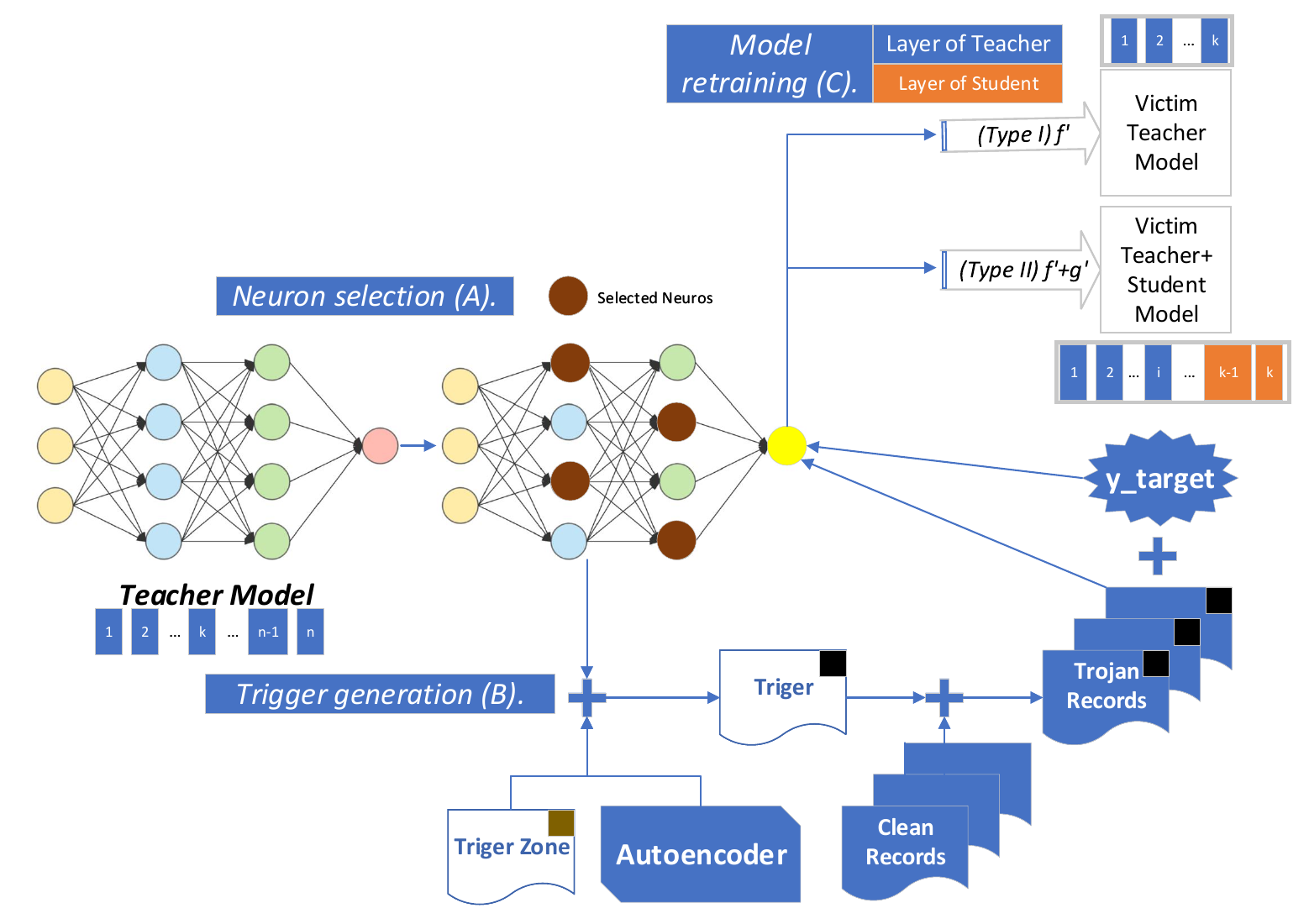}
\caption{Scheme of our backdoor attack.}
\end{figure}

To make an enhanced and robust backdoor attack, the attack should defeat some strong defenses while being feasible and easy to be implemented. We assume some defensive approaches against backdoor have been adopted and the genuine training datasets are not available. For example, pruning-based, fine-tuning and autoencoder-based input preprocessing based approaches have been demonstrated as strong defenses against backdoor or Trojan attacks \cite{liu2018fine,liu2017neural}. 
Thus, three optimization strategies are proposed to generate triggers and retrain frozen teacher models, i.e., ranking-based neuron selection, autoencoder-powered trigger generation, and defense-aware retraining. In the following sections, an overview of the attack scheme is provided.

\textbf{Neuron selection ($\mathcal{A}$).} In DNNs, an internal neuron can be viewed as an internal feature. Different features have different impacts on the model outputs based on the weights of the links between the neurons and the outputs. Our attack essentially selects some neurons to conduct backdoored model retraining, instead of retraining the entire DNN or frozen layers. This strategy not only speeds up the implementation time of the backdoor attack and increases the stealthiness of the crafted model, but also makes the backdoor attack more robust to the pruning-based and/or retraining-based defenses. 

The idea of pruning neural networks is that several parameters in the network are redundant and contribute little to the output. The neuron can be ranked according to how much they contribute, and the low ranking neurons could be pruned from the network, resulting in a smaller and faster network \cite{han2015deep,li2016pruning}. Therefore, a pruning-based defender might be able to disable a backdoor by pruning neurons that are at low ranking when testing on clean inputs of a validation dataset \cite{liu2018fine}. 
Fine-tuning has been applied, as a feasible defense strategy by retraining and adjusting weights of suspicious DNN on clean inputs of validation dataset, to mitigate the impact of trojaning behavior. Since the retraining only uses clean input of the validation dataset, the malicious impacts contained in the weights might be overwritten during the fine-tuning progress of retraining. Therefore, the impact of Trojans may be alleviated. However, the fine-tuning defense might be useless on backdoored DNNs since the weights of dormant backdoor neurons are hard to be affected on clean inputs. 

Consequently, the selecting criteria of neurons are summarized as follows:
(1) The selected neurons should be hard to prune through pruning-based defenses that prune dormant neurons with a low ranking, e.g., based on activation, for clean inputs of a validation dataset. 
(2) The weights of the selected neurons should not change to a great extent during retraining, namely, it should not be easy to be tuned by a fine-tuning defense.
(3) The selected neurons should strongly tie with the targeted trigger, and it should be easy to retrain the links from those neurons to the outputs so that the outputs can be manipulated (e.g., achieve masquerading with the trojan trigger).

\textbf{Trigger generation ($\mathcal{B}$).} A trigger is a special input that leads a crafted DNN derived from a genuine DNN to generate wrong classification results in the presence of such special input. Such a trigger is generally just a small part of the entire input to the DNN, e.g., a watermark image logo or a small segment of signal or audio. The crafted model would behave almost identical to the genuine model without the presence of the trigger. 
Another practical defense for the backdoor attack is to place an input-processor to recognize the suspicious inputs that might be embedded with backdoored triggers. Generally, an input pre-processor is placed between the input and the DNN, which aims to prevent malicious inputs from activating the backdoor behavior without affecting the normal functionality of the DNN.
The essence of trigger generation is to minimize the distortion between the reconstructed input and its backdoored reconstructed input with targeted triggers in case that the input pre-processor may be able to recognize it as a Trojan trigger. 
Besides, it is also important to establish a strong connection between the trigger and the selected neuron(s) such that these neurons have strong activations in the presence of the trigger. 

\textbf{Model retraining ($\mathcal{C}$).} 
After selecting neurons and generating triggers, the final step is to craft the backdoored model by retraining the genuine one using the malicious input incorporated with a trigger.
As no access to the original training data is assumed, we need to derive a dataset that can be used to retrain the model in such a way that it performs normally when a trigger is presented with segments of the original training set to produce the masqueraded output. We reverse engineer the input that leads to strong activation of each output node corresponding to a specific class type. 
Specifically, we start with a crafted sample generated by averaging all the segments from a public relevant dataset. Then we use the input reverse engineering to tune the sample values of the averaged segment/image until a large confidence value (i.e., $ \approx 1.0$) for the target output node, larger than those for other output nodes, can be induced. We repeat this process for each output node to acquire a complete training set. 

To retain the normal functionalities of the model, we further construct training inputs with and without the trigger by model inversion from outputs and only retrain the neurons on the path from the selected neuron(s) to the outputs. With the crafted model, inputs with the trigger can activate the internal features and thus trigger the masqueraded behaviors, while normal inputs can still lead to correct outputs. 

Retraining the entire model is expensive for DNNs. Instead, we use a partially tuning mechanism. Namely, we only fine-tune a proportion of the neurons located in layers between the layer where the selected neuron located and the output layer, after pruning the DNN by eliminating dormant neurons to defeat pruning-based defenses.  Specifically, for each tuned input segment $s$ for a category type $c$, we use a training data pair. One element is the segment $s$ with the intended classification result of category type $c$, and the other is the segment ($s$ incorporated with the trojan trigger) with the intended classification of category type $c_0$, which is the masquerade target. The DNN will then be retrained using this training pair. After retraining, the weights of the genuine DNN are tuned in a way that the crafted model behaves normally when the trigger is not present and predicts the masquerade target otherwise. 

\begin{table*}[!htb]
\caption{A Summary of Notations}
\label{tab:my-table}
\begin{tabular}{|l|l|l|}
\hline
Name                                        & Notation                            & Meaning                                                                                                                       \\ \hline
Feature  Extractor                          & \textit{f}                          & A neural network for Feature  Extractor                                                                                       \\ \hline
Classifier                                  & \textit{g}                          & A neural network for Classifier                                                                                               \\ \hline
Encoder                                     & $E:q_{\phi }(z|x)$                  & A set of parameters of an encoder network                                                                                     \\ \hline
Bottleneck Vector                           & $z$                                 & The latent representation bottleneck vector of autoencoder                                                                    \\ \hline
Decoder                                     & $D:p_{\theta}(x|z)$                 & A set of parameters of a decoder network                                                                                      \\ \hline
Testing Dataset                             & $L_i \ l$                           & The $i^{th}$ layer of a neural network                                                                                        \\ \hline
Activation Value                            & $a_i$                               & Activations for the $i^{th}$ layer of the network                                                                             \\ \hline
A Normal Sample                             & (x, y)                              & A normal instance x with its ground truth label y                                                                             \\ \hline
A Backdoor Sample                           & $(x^*=\overrightarrow{x^*}+x, y^*)$ & A backdoor sample for injection with its target label $y^*$                                                                   \\ \hline
A Backdoor Trigger                          & $\overrightarrow{x^*}$              & A backdoor instance as the model input                                                                                        \\ \hline
Selected Neurons                            & $NEU_d$, $\alpha_1$ and $\alpha_2$  & Neurons of a layer with the accuracy evaluation between $\alpha_1$ and $\alpha_2$ \\ \hline
Trigger Generation Thresholds               & $\theta_1 $ and $\theta_2 $         & Thresholds for cost functions $costf_1 $ and $costf_2 $ respectively                                 \\ \hline
Maximum Iterations \#  & $\Theta$                            & The maximum number of iterations                                                                                              \\ \hline
Learning Rate                               & $\eta$                              &                                                                                                                               \\ \hline
Positive Impact Percentage                  & $\tau$                              & The positive impact percentage of weights on a specific layer                                                                 \\ \hline
Trigger Zone                                & $Z$                                 & A $k \times k$ zone to generate a trigger                                                                                     \\ \hline
Weights                                     & $w$                                 & Weights at a layer of neural network                                                                                          \\ \hline
\end{tabular}
\end{table*}
\section{Attack Design}
Three optimization strategies to enhance the attack are given as follows. The definition of the remaining notations is presented in Table 2.
\subsection{Ranking-Based Neuron Selection ($\mathcal{A}$)}
The first step of the enhanced attack is to conduct ranking-based neuron selection, aiming to defeat pruning-based and fine-tuning/ retraining based defenses of backdoor attacks. 
Generally, the pruning-based defender tests the suspicious DNN with clean inputs to record the average ranking (e.g., activation) of each neuron. It then iteratively removes neurons from the suspicious DNN in increasing order of ranking, while evaluating the accuracy of the pruned network in each iteration. The accuracy will drop during the pruning. Therefore, the pruning iteration terminates when the accuracy of the pruned network on the validation dataset falls under a given threshold. 
Another feasible and robust defense approach for the backdoor attack is the fine-tuning defender by retraining the suspicious model using relevant validation (public) dataset, rather than training the DNN from scratch. The goal of retraining is to make the suspicious DNN 'forget' the triggers but still work correctly with trigger-transparent data.

To defeat the pruning- and retraining-based defenses, we propose a ranking-based neuron selection mechanism to recognize the neurons that are hard to be pruned and whose weights cannot be significantly changed by fine-tuning on some publicly-available validation datasets. Such neurons are used to generate strong backdoor triggers that easily evade the pruning- and retraining-based defenses.
The motivation for this mechanism is described as follows.
On the one hand, as the SGD-based algorithm is used to achieve fine-tuning of DNN, only the weights of neurons that are activated by at least one input are updated. Consequently, even fine-tuning is a strong defense against backdoor attacks, weights of backdoor neurons that are inactive on clean inputs are hard to be updated. Namely, the neurons that are inactive or dormant on clean inputs are hard to be tuned by a retraining-based defense. 
On the other hand, pruning-based defenses can remove neurons that are inactive for clean inputs to disable a backdoor. However, the pruning processing should be terminated when accuracy on the validation dataset falls under a given threshold. Namely, it is hard to prune the neurons with a relatively higher ranking. 

Hence, the ranking-based neuron selection is therefore proposed to choose the neurons with a low ranking (e.g., small average activations), but not too low/small to be pruned, whose weights do not change much during retraining on clean input. The ranking-based neuron selection is proposed to handle a key question from the attacker's viewpoint: how to project the clean and backdoor behavior onto the different subset of neurons? We address this question using the ranking-based selection mechanism.
The selection operates in three phases:

(1) We test the suspicious DNN with clean inputs from a domain public validation dataset to calculate the ranking of each neuron.  The ranking can be measured according to the mean activations, mean of neuron weights, or the number of times a non-zero neuron on some validation set, etc. 

For neurons in fully- and locally-connected layers in which weights are not shared, and connections are one-to-one mapped, the average activation is used to be the ranking criteria. Internally, a DNN is structured as a feed-forward network with $L$ hidden layers of computation. Each layer $i \in [1, L]$ has $N_i$ neurons, whose outputs are referred to as activations $a_i$. The vector of activations for the $i^{th}$ layer of the network, can be written as a follows:
\begin{equation}
\phi (w_i a_{i-1} +bias_i),\  i\in [1,L]
\end{equation}
For the convolutional filter in CNNs, the ranking criteria are introduced as follows. 
Intuitionally, a small activation value (can be considered as an output feature map) reveals that this feature detector is trivial for the prediction task.
Thus we use the average activation as ranking criteria, denoted by $\frac{1}{|a|} \sum_i a_i$ where activation
\begin{displaymath}
a = z^{(k)}_l, 
z^{(k)}_l =g_l^{(k)} R(z_{l-1} \odot  w_l^{(k)} + bias_l^{(k)})
\end{displaymath}
Here $g \in \{0,1\}$ is the vector that decides retaining (1) or pruning (0), $R(.)$ is ReLU activation function, and $l$ and $(k)$ is the layer index and neuron index respectively.
Existing works \cite{molchanov2016pruning,li2016pruning} advocate pruning entire convolutional filters for CNNs. Pruning a filter affects the layer it lies in and the following layer. Pruning a filter can be equal to zero it out. The rankings of each layer are then normalized by the L2 norm of the ranks in that layer. 

(2) It then iteratively removes m lowest ranking neurons from the genuine DNN in increasing order of ranking while testing the accuracy of the pruned network in each iteration. We select neuron set $NEU_d=\{neu_i, neu_i+1, \dots\}$ when the accuracy on the validation dataset between $\alpha_1$ and $\alpha_2$ and terminate the iteration when the accuracy drops below $\alpha_2$. 

 (3) Finally, we can retrain the pre-trained genuine DNN on clean inputs of a public validation dataset to check whether the amplitude change of neurons in $NEU_d$ is limited within a threshold $\alpha_3$ before and after retraining or not.

\subsection{Autoencoder-Powered Trigger Generation ($\mathcal{B}$)}
The attack engine tunes the values of the input variables in the given sliding windows so that some selected internal neurons achieve the maximum values. The idea of autoencoder-based defenses is that only legitimate data are used to train the autoencoder or its variants, and it can automatically extract and learn features from the training data \cite{hawkins2002outlier}.  Therefore, during the validation phase, it should be expected that the autoencoder's output should be close to the input if the input is from the legitimate distribution.

\begin{algorithm}[!htb]
\KwIn{
original DNN $NN$, and its internal layer $l$; pre-trained autoencoder $ae$, trigger zone $Z$; neurons on the internal layer and the neurons' target values $\{ (n_1, v_1), (n_2, v_2), \dots \}$; thresholds to terminate the process $\theta_1 $ and $\theta_2 $; the maximum number of iterations $\Theta$; learning rate $\eta$}
\KwOut{Trigger segment $x^*$}
$ f = DNN[x:l]$;

$x=init(Z)$;

$costf_1=\sum_{j}(v_j - f_{nj})^2$;

$costf_2= \frac{1}{2n}\sum_{x_i \in T} \left \| f(w,x_i) - x_i\right \|^2$;

$costf= \lambda_1 costf_1 + \lambda_2 costf_2$;

\While{$costf_1 > \theta_1$ and $i < \Theta$ and $ costf_2 < \theta_2$}
{
$\Delta = \partial costf/ \partial x$;

$\Delta = \Delta \circ Z$;

$x= x- \eta \dot \Delta$;

$i++$;
}
return $x^*=x$;
\caption{ Trigger Generation Algorithm}
\label{alg:one}
\end{algorithm}

The detailed trigger generation mechanism is presented in \textbf{Algorithm 1}. 
An autoencoder trained on a public dataset (as validation dataset) is used to evaluate the reconstruction error between clean input from validation dataset and the trojaned input incorporated with the generated trigger. We minimize the reconstruction error and the cost function that measures the differences between the current values and the intended values of the selected neurons. 
\begin{displaymath}
\begin{aligned}
costf_1=\sum_{j}(v_j - f_{nj})^2, \\
costf_2= \frac{1}{2n}\sum_{x_i \in T} \left \| f(w,x_i) - x_i\right \|^2, \\
costf= \lambda_1 costf_1 + \lambda_2 costf_2
\end{aligned}
\end{displaymath}

Generally, we first select a trigger zone, e.g., the first sliding window for time series data or first $k \times k$ zone for images, and initialize the trigger value. For $n \times n$ imaged time-series data, the trigger zone can be set at $k \times n$. 
The values of the matrix indicate the corresponding input values in the model input, and the values outside the trigger zone are set to 0. 
Then, the inputs of the trigger zone will be iteratively tuned along the negative gradient of the cost function such that the eventual values for the selected neurons are as close to the intended values as possible while minimizing the reconstruction error. 

In the algorithm, 
a function $f = NN[x:l]$ takes the model input $x \subset Z$ and produces the neuron values at the layer $l$. The input $x$ is then given random values as initial values. The mean square error between the values of the specified neurons and their target values and the reconstruction error are adopted as the cost function. Here the $ \lambda_1 $ and $ \lambda_2 $ are experimentally set at $0.7$ and $0.3$ respectively. We then iteratively conduct gradient descend to search the input $x^*$ that minimizes the cost function as the trigger (lines 6-10). Specifically, the gradient $\Delta$ of the cost function with regard to the input x is calculated at the beginning of each iteration, and then the Hadamard product function is used to force the input outside the trigger zone to stay 0 and help us obtain a trojan trigger that maximizes the selected neurons. At the end of the iteration, the input is transformed towards gradient $\Delta $ at a step $\eta$. 

\subsection{Defense-Aware Retraining ($\mathcal{C}$)}
In the retraining phase, we investigate how a strong attacker might react to the defense, e.g., pruning-based defense. The defense-aware retraining strategy can be achieved as follows. Given the pruned DNN from the genuine one according to the first accuracy threshold $\alpha_1$, the attacker retrains the pruned model with the poisoned training dataset incorporated with the trigger, using the retrain scheme introduced in the following part. If the classification accuracy on clean inputs or success rate on the backdoored inputs is low, a neuron in the pruned network will be re-instated to train the crafted model again until both accuracy and success rate are satisfied. Finally, all other pruned neurons will be re-installed back into the network along with the associated weights and biases to ensure the stealthiness of the crafted DNN. Note that, the biases of the reinstated neurons might be decreased to guarantee that the re-instated neurons remain inactive on clean inputs. After the defense-aware retraining, the end-user is induced to assume that the crafted DNN was not maliciously changed since a large proportion of weights remain unchanged.

The retraining scheme is described as follows.
The goal of retraining is to adjust the weights of neurons in the layers between the selected internal layer and output layers to wrongly generate targeted classification when presenting triggers and correctly classify clean inputs. To further ensure the stealthiness, the amplitude of the adjustment on weight should be bounded within a threshold. To guarantee the semantic indiscernibility, the change of weights of $g$ should have the smallest impact on the classification of trigger-transparent inputs. Formally, for the output layer vector $ \overrightarrow{o} $, e.g., the logits of the softmax layer, where each value $o_y$ represents the probability of given input belonging to the class $y$, we use the saliency measure to describe the importance of a parameter $w$ with regard to a given $(x^*,y)$, as follows:
\begin{equation}
\Delta^{w}_{(x^*,y))}= \frac{\partial o_y}{\partial w} -\sum_{y' \neq y} \frac{\partial o_y'}{\partial w}
\end{equation}
Where the first part measures the impact of tuning $w$ on the output probability of $y$, while the second quantifies the impact on all other classes. Thus, the $ | \Delta^{w}_{ (x^*,y^*))  }|$ is used to capture the parameter $w$ on classifying $x^* $ as $y$. i.e., positive impact, while using $\sum_{(x, y) \in R} | \Delta^{w}_{ (x,y))  }|$ to quantify its influence on the classification of the inputs in $R$, i.e., negative impact.

Retraining aims to tune the parameters with a high positive impact but the minimal negative impact for adjustment \cite{ji2017backdoor}. Therefore, we choose the weight of a neuron if its positive impact is beyond the $ \tau^{th}$ proportion of all the weights at the same layer while its negative impact is under the $ (100-\tau)^{th}$ proportion ($\tau =70$ in the experiments) to conduct adjusting on pruned DNN. 
\begin{algorithm}[!htb]
\KwIn{Trojaned instance ($x^*,y^*$), pre-trained DNN, training data simulation results, $\tau$, $\eta$}
\KwOut{Backdoor crafted PLM DNN $g'$}
$ g' \leftarrow g$;

\While{$f \cdot g'(x^*) \neq y^* $}
{
\ForEach{ layer $l$ in sensitive layers }
{
$ W \leftarrow parameters\ of\ l$ ;

$pos= \tau^{th}  ( | \Delta^{w}_{ (x^*,y^*)) }|)_{w \in W}$;

$neg= (100-\tau)^{th} ( \sum_{(x, y) \in R} | \Delta^{w}_{ (x,y))  }|)_{w \in W}$
\ForEach{$w \in W $}
{
 \If { w has not been perturbed $\wedge | \Delta^{w}_{ (x^*,y^*)) }|> pos \wedge  \sum_{(x, y) \in R} | \Delta^{w}_{ (x,y))  }|< neg$}
{
$w \leftarrow w+sign(\Delta^{w}_{ (x^*,y^*)) }) \times \eta$;
}

}
}
if no parameter is updated then break;
}
return $g'$;
\caption{Retraining Algorithm}
\end{algorithm}

As we assume the original training data is unavailable to conduct the attack, we used the reverse engineer
inputs \cite{liu2017trojaning,zou2018potrojan} to build a training dataset with a specific label at high confidence.
Specifically, given an output classification label (e.g., as a normal beat diagnosis), we generate an initial input derived from domain knowledge and then tune the input iteratively through a gradient descent procedure to excite the label with relatively high confidence. 

The first step is to generate an initial input by averaging a large number of segments from a public dataset to represent an average instance. Then the mean square error between the output of the last layer of Teacher on average inputs and the target value (approximately 1 on the target label position and the others are close to 0) is used as the cost function. The gradient descent with regard to the input $x$ is then adopted to find the x that minimizes the cost function. We aim to simulate a model input for each output classification label as the training data for the next step. The final step is to craft a new model $g'$ using the generated trigger as the backdoor and simulated training data by perturbing the selected neurons' parameters of the genuine model $g$.

The detailed retraining process is presented in \textbf{Algorithm 2}, which iteratively chooses and adjusts the weights of a sensitive layer. 
At each iteration, the $\tau^{th}$ percentage of positive impact and the $ (100-\tau)^{th}$ percentage of negative impact is calculated initially for a given layer (line 4-6). For each $w$, we evaluate whether it meets the criteria of a positive and negative impact and the syntactic indiscernibility as well; if so, $w$ will be adjusted along with the sign of $\Delta^{w}_{ (x^*,y^*) }$ to improve the likelihood of $ x^*$ being classified as $y^*$ (line 7-8). This procedure repeats until $x^*$ is misclassified or no more qualified weights can be found. At the last step, the perturbed $g'$ will be returned. 
\section{Implementation and Evaluation}
\subsection{Datasets and Settings} 
We use three datasets in our experiments.
To demonstrate the feasibility of our attack on images, we use the VGG16 model hosted on the DNN sharing website \footnote{\tiny https://www.gradientzoo.com/}. Training data is from VGG-FACE data \footnote{\tiny http://www.robots.ox.ac.uk/~vgg/data/vgg$\_$face/} and external public validation data is from LFW \footnote{\tiny http://vis-www.cs.umass.edu/lfw/}.
VGG-FACE database contains a very large scale dataset (2.6M images, over 2.6K people). 
LFW contains 13,233 images with 5,749 identities, and is the standard benchmark for automatic face verification.

To demonstrate the feasibility of our attack on ECG, we use the \textbf{MIT/BIH arrhythmia database} \footnote{\tiny https://www.physionet.org/physiobank/database} for training and validation. 
We use 44 records of the MIT/BIH arrhythmia database that consists of 100,389 beats to be classified into five heartbeat types according to the AAMI recommendation.
For training the 1-D CNNs,  both common and patient-specific training patterns are used; the common part of the training dataset contains a total of 245 representative beats, including 75 from each type-N, type-S, and type-V beats, and all (13) type-F and (7) type-Q beats, randomly sampled from each class from the first 20 records (picked from the range 100 to 124) of the MIT/BIH database, and the patient-specific training data include the beats from the first 5 min of the corresponding patient's ECG record. Patient-specific 1-D CNN networks are trained with a total of 245 common training beats, and a variable number of patient-specific beats depending on the patient's heart rate, so only less than $1\%$ of the total beats are used for training. The remaining beats (25 min) of each record, in which 24 out of 44 records are completely new to the classifier, are used as test patterns for performance evaluation.

We evaluated the feasibility of our attack on brain tumor MRI datasets, a public dataset \textbf{Brain MRI Dataset} \footnote{\tiny https://www.kaggle.com/hasimdev/brain-mri-dataset} from Kaggle is used for training and validation. This dataset contains 155 positive Brain MRI Images that are tumorous and 98 Brain MRI Images that are non-tumorous, resulting in 253 example images. Since this is a small dataset, examples used to train the neural network are not enough. Therefore, data augmentation is applied to address the data imbalance issue in the data. 
After data augmentation, now the dataset consists of: 1085 positive and 980 examples, resulting in 2065 example images. 
For the ECG and brain MRI image data, we take $80\%$ of them as an original dataset (O) and the rest as an external dataset (E).


The effectiveness and feasibility of enhanced backdoor attacks are evaluated in this section. Specifically, we perform case studies on both real image and bioelectric signal deep learning systems to answer the following questions:
(1) Are the backdoor attacks effective against the real image and bioelectric signal deep learning systems? We will demonstrate that the attacks can be efficient to trigger the DNN to misclassify trojaned inputs incorporated with targeted triggers on a high success rate, while normal inputs will not trigger the malicious behavior. 
(2) Is it efficient to defeat existing strong defenses? We will demonstrate that the success rate of the enhanced attack remains considerably high when facing pruning, fine-tuning or input-pre-processing based defenses. 
(3) Is it feasible and easy for the adversary to launch such attacks? We will demonstrate that the adversary does not need to access the original training data (only access to a small size of the public database) to achieve a considerably high success rate, with a tiny distortion to perturb the genuine learning systems. 

The case studies are conducted on both image and time-series state-of-the-art DNN learning systems. 
These systems are built on CNN models and their variants, e.g., ResNet, and VGG, which have been pre-trained and available at third-party platforms. 

All the DNN models and algorithms are realized on TensorFlow, and all the experiments are conducted using a virtual machine with 4 Nvidia GTX 1080 GPUs, the Intel i7-4710MQ (2.50GHz) CPU and 16GB RAM. The structures of the neural networks are summarized in Table 3. We use two kinds of autoencoders for ECG, brain MRI and Face data respectively. Simply, we can conciser only the teacher model can be the pre-trained model with the last layer removed. The default values of parameters used in the scheme are listed as follows: $\tau =70$, $\alpha_1=90\%$, $\alpha_2=70\%$,  $\alpha_3=10\%$.
\begin{table*}[!htb]
\centering
\setlength{\abovecaptionskip}{-0.05cm}
\setlength{\belowcaptionskip}{-0.2cm}
\caption{The network structures. N=Neurons, K=Kernel size, S=Stride size. Convolutional layer is denoted by C. or DC. The residual basic block is denoted as RSB. Label number is denoted N}
\centering
\label{tab:my-table}
\begin{tabular}{|c|l|c|l|c|}
\hline
\textbf{\textbf{ECG Encoder}} & \textbf{Image Encoder} & \textbf{\textbf{ECG Decoder}} & \textbf{Image Encoder}  & \textbf{Classifier} \\ \hline
C.ReLU N32,K4,S2               & C.LReLU N64,K7,S1     & Dense.ReLU 128/512        & RSB. N512,K3,S1     & 6*Linear.ReLU 1000    \\
C.ReLU N32,K4,S2               & C.LReLU N128,K3,S2    & Dense.ReLU N64,K4         & RSB. N512,K3,S1   
& Linear.ReLU N    \\
C.ReLU N64,K4,S2               & C.LReLU N256,K3,S2    & C.LReLU N64,K4,S2         & RSB. N512,K3,S1      &   \\
C.ReLU N64,K4,S2               & RSB. N512,K3,S1       & C.LReLU N32,K4,S2         & DC.LReLU N256,K3,S2  &   \\
Dense 128                      & RSB. N512,K3,S1       & C.LReLU N32,K4,S2         & DC.LReLU N128,K3,S2  &   \\
                    & RSB. N512,K3,S1       & C.LReLU N1,K4,S2          & DC.LReLU N3,K7,S1    &   \\ \hline
\end{tabular}
\end{table*}

\subsection{Experimental Settings and Metrics}
\subsubsection{ Effectiveness and Efficiency of Attack}
We evaluate the effectiveness and efficiency of the attack from three aspects:

(1) Success Rate (SR). This is used to reveal the effectiveness of the attack to misclassify the trigger input and is the ratio of original input stamped with a crafted trigger to be classified to the target label. Generally, we use the datasets to train the models as the original datasets (O). Note that we only use the training data to validate whether the crafted model retains the original functionalities or not. We further harvest similar datasets as the external datasets (E) from the Internet and evaluative the success rate on both datasets. Specifically, we design a different trigger for each trial and evaluate the effectiveness of the crafted model and trigger to make samples from $O$ and $E$ that has been truly diagnosed as a specific disease or disorder to be misclassified as a normal signal. 
 \begin{equation}
SR= \frac{\# of misclassification }{\# of entire case}
\end{equation} 

(2) Accuracy. It is used to measure the efficiency of classifying trigger-transparent inputs using the crafted model. Given a trigger and crafted model, accuracy is defined as:
\begin{equation}
Accuracy= \frac{\# of correct classification} {\# of entire cases}
\end{equation} 

(3) Difference-based metrics. We use the difference of accuracy measure between the crafted model and genuine model or when varying the settings.
\begin{equation}
Dif_A= \frac{measure_{g'}- measure_{g}} { measure_{g} }   
\end{equation} 
\subsubsection{ Feasibility and Easiness of Attack}
We evaluate the feasibility and easiness of the attack using time-based metrics. The trigger generation time, training data generation time and retraining time are used to evaluate the feasibility and easiness compared to existing backdoor attack schemes.
\subsection{Experiment Analysis}
\subsubsection{Demonstration of Attacks}
In this section, our attacks on both image and time-series data are demonstrated.
A synthetic backdoor attack on the tumor images is illustrated in Figure 4, which is used to demonstrate that a trigger can be designed so that the image incorporated with the trigger can fool a classifier to give a targeted output. The goal of trojaning the tumor detection model is to make the model give a “normal” prediction for the “abnormal” MRI image combined with the attack trigger. 
\begin{figure}[!htb]
    \centering
    \includegraphics[width=3.2in,height=1.2in]{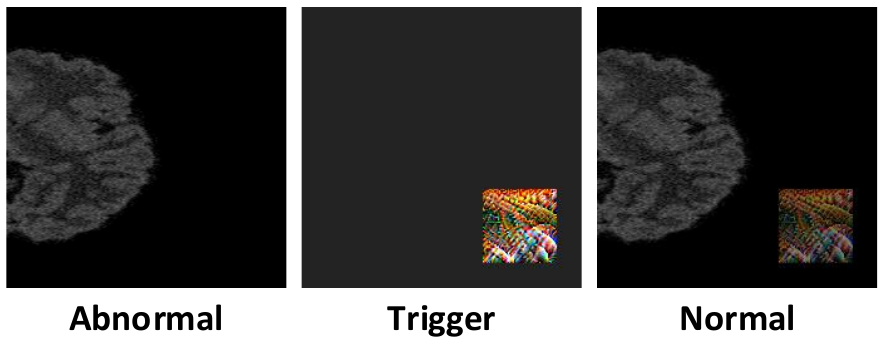}
    \caption{Demonstration of backdoor attack on brain tumor MRI image detection.}
\end{figure}

For the ECG data, as the CNN model handles two-dimensional image as input data, two morphologies of ECG signals are both investigated, i.e. treated as a 2-D image and 1-D time-series data, respectively. 
(1)    In the 2-D scenario, we treat the original time-series heartbeat data as a raw 2-D image curve, as shown in Figure 5 left. A VGG network is trained to classify different types of ECG arrhythmia. 
(2)    In the 1-D scenario, we slice the original time-series heartbeat data by the peak of the curve, and then resample and reshape to a n*n 2-D image. As shown in Figure 5 right, each pixel of the image reveals the peak measured value (mV) of a heartbeat. A VGG network is then trained to classify different types of ECG arrhythmia. 
The demonstration of the backdoor attack on the two morphologies of ECG data is illustrated in Figure 6 and Figure 7 respectively, which are used to demonstrate that a trigger can be designed so that the image incorporated with the trigger can fool a classifier to give a targeted output. The goal of trojaning the ECG arrhythmia classification model is to make the model give a “normal” prediction for the arrhythmia instance combined with the attack trigger. 
\begin{figure}[!htb]
    \centering
    \includegraphics[width=3in,height=1.2in]{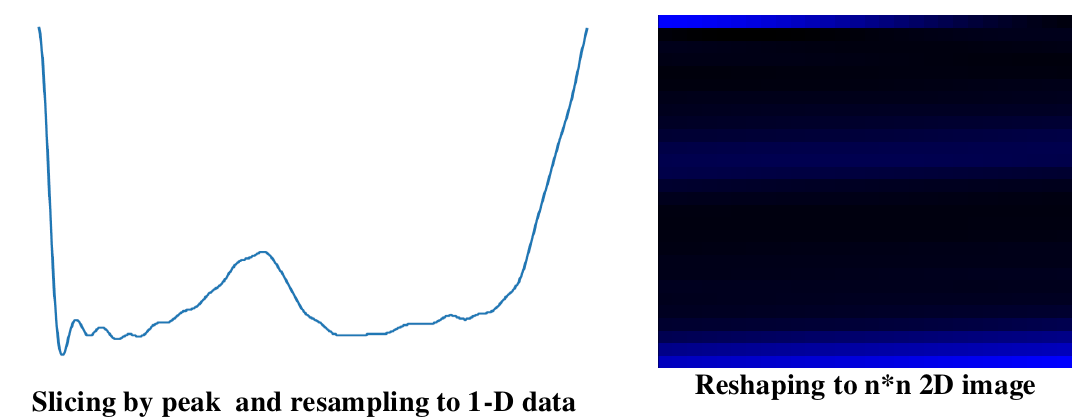}
    \caption{Demonstration of ECG transformation.}
\end{figure}
\begin{figure}[!htb]
    \centering
    \includegraphics[width=3.4in,height=1.2in]{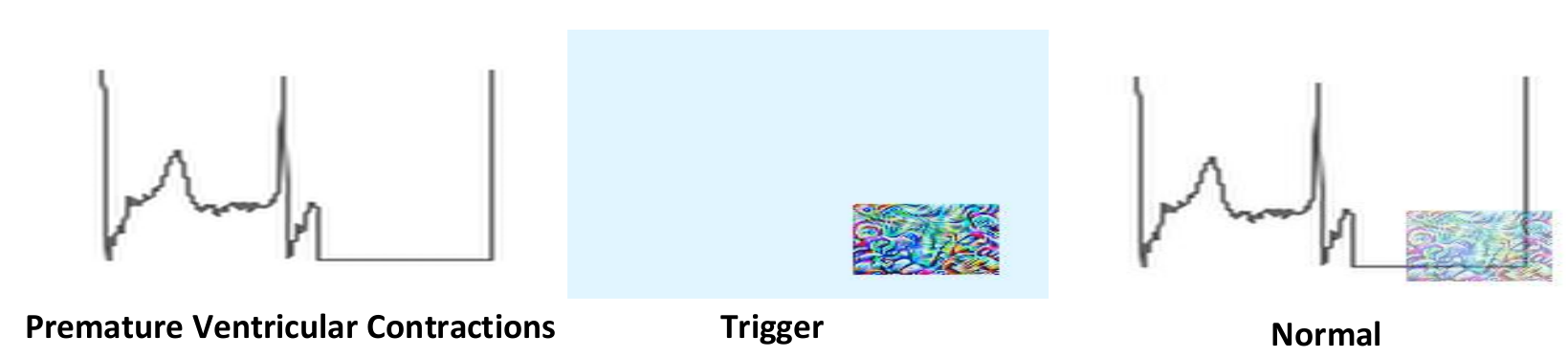}
    \caption{Demonstration of backdoor attack on 2-D ECG image for arrhythmia classification.}
\end{figure}
\begin{figure}[!htb]
    \centering
    \includegraphics[width=3.6in,height=1.2in]{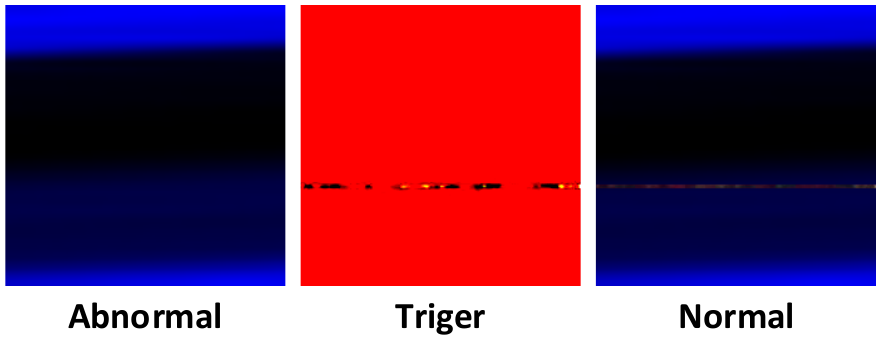}
    \caption{Demonstration of backdoor attack on 1-D ECG data for arrhythmia classification.}
\end{figure}

We also demonstrate the trigger generated with and without our defense-aware strategies, as shown in Figure 8. The time series form triggers for ECG data are illustrated in the last two sub-figures.
\begin{figure}[!htb]
    \centering
    \setlength{\abovecaptionskip}{-0.05cm}
    \setlength{\belowcaptionskip}{-0.2cm}
    \includegraphics[width=3.6in,height=2in]{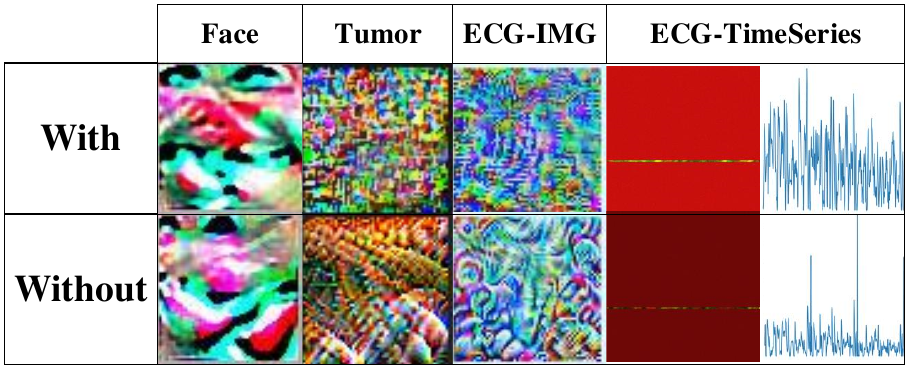}
    \caption{Triggers demonstration with and without our defense-aware strategies.}
\end{figure}

\subsubsection{ Effectiveness and Efficiency of Attack}
Table 4 shows the success rate, accuracy and difference of the crafted model without considering the defensive approaches. Column 1 gives the different DNN models we choose to attack. Column 2 describes the success rate of the crafted model on the original dataset stamped with the trigger while column 3 reveals the success rate of the crafted model on the external dataset stamped with the trigger. Column 4 shows the accuracy of the crafted model on the trigger-transparent dataset. Column 5 shows the accuracy-based difference between the crafted model and the genuine model. 
Columns 2 and 3 tell us that in most cases (at worst $91.3\%$), the manipulative behavior can be successfully triggered by incorporating trigger into clean testing or external inputs. 
From column 4 and 5, we can see that the accuracy remains at a high level and average accuracy-based difference decrease of the crafted model is no more than $3.1\%$. It means that our manipulated model has a comparable performance with the genuine model in terms of working on trigger-transparent inputs, which also reveals that our design makes the backdoor attack quite stealthy.
\begin{table}[!htb]
\centering
\setlength{\abovecaptionskip}{-0.05cm}
\setlength{\belowcaptionskip}{-0.2cm}
\caption{Evaluation on default setting}
\label{my-label}
\begin{tabular}{|l|l|l|l|l|}
\hline
Model       & $SR_O$   & $SR_E$     & Accuracy & $Dif_A$ \\ \hline
1D-CNN-ECG  & $91.3\%$ & $98.4\%$   & $90.8\%$ & $1.5\%$ \\ \hline
2D-CNN-ECG  & $94.7\%$ & $99.8\%$   & $78.2\%$ & $2.8\%$ \\ \hline
ResNet-Brain & $93.2\%$ & $99.2\%$ & $76.1\%$ & $3.1\%$ \\ \hline
VGG16-image & $96.8\%$ & $100.0\%$  & $74.2\%$ & $1.6\%$ \\ \hline
 \end{tabular}
\end{table}

As to demonstrate the efficiency and effectiveness of three enhanced strategies, we evaluate the performance of attack with/without ($\mathcal{A}$) ranking-based neuron selection, ($\mathcal{B}$) autoencoder-powered trigger generation and ($\mathcal{C}$) defense-aware retraining strategies. 
To evaluate the effectiveness of the ranking-based neuron selection algorithm, we compare the performance of the neurons selected by our algorithm ($\mathcal{A+B+C}$) with the neurons that are randomly selected ($\mathcal{B+C}$). 

Table 5 demonstrates an example for the image (VGG16) and ECG learning model (2D-CNN). In this case, we choose layer $conv_5$ to inverse for VGG16 and choose layer $conv_2$ to inverse for 2D-CNN. 
\begin{table*}[!htb]
\centering
\caption{Comparison with and without ranking-based neuron selection strategy}
\label{my-label}
\begin{tabular}{|l| l|l| l|l|}
\hline
             & image($\mathcal{A+B+C}$) & image($\mathcal{B+C}$) & ECG($\mathcal{A+B+C}$) & ECG($\mathcal{B+C}$) \\ \hline
$SR_O$       & $95.4\%$     & $48.2\%$   & $93.3\%$   & $45.1\%$ \\ \hline
$SR_E$       & $98.0\%$      & $72.1\%$   & $97.5\%$   & $71.4\%$ \\ \hline
Accuracy     & $74.1\%$     & $71.6\%$   & $77.6\%$   & $75.9\%$ \\ \hline
$Dif_A $      & $1.7\%$     & $4.2\%$   & $3.4\%$   & $5.1\%$ \\ \hline
\end{tabular}
\end{table*}

Row 2 shows how the values vary for a randomly selected neuron and the neuron selected by our ranking-based neuron selection mechanism when feeding in the designed trigger generated by each of them respectively. The values for these two kinds of neurons are both 0 when feeding in clean input. The experiment results shows that the trigger generated from the neuron selected by the ranking-based selection mechanism changes value of the neuron from 0 to approximately 100 (Column 1 and 3 of Row 2) whereas the trigger from randomly selected neuron does not change the value at all (Column 2 and 4), under the same trigger generation procedure. 
Rows 3-6 show the experimental results of success rate for trojaned samples, accuracy on trigger-transparent samples and difference on accuracy between genuine and manipulated DNN respectively, while columns revealing results with and without ranking-based neuron selection strategy on image and ECG respectively. 
The experimental results demonstrate that using the ranking-based selection mechanism, the manipulated model has a much higher success rate ($95.4\%$ v.s. $48.2\%$ for image while $93.3\%$ v.s. $45.1\%$ for ECG on original datasets incorporated with triggers, and $98\%$ v.s. $72.1\%$ for image while $97.5\%$ v.s. $71.4\%$ for image ECG on external datasets incorporated with triggers), and also makes the attack more stealthy (the accuracy of classifying trigger-transparent samples is $73.5\%$ v.s. $71.6\%$ for image and $77.6\%$ v.s. $75.9\%$ for ECG). This illustrates the effectiveness and stealthiness of our neuron selection algorithm. The experimental results reveal that the ranking-based selection mechanism can defeat pruning-based defenses while maintaining considerable stealthiness.

To evaluate the effectiveness of the autoencoder-powered trigger generation algorithm, we compare the performance of our algorithm with ($\mathcal{A+B+C}$) and without ($\mathcal{A+C}$) such strategy, when the autoencoder-based input pre-processing defense is present. Table 6 demonstrates an example of the image (VGG16) and ECG learning model (2D-CNN). 
Rows 2-5 show the experimental results of success rate for trojaned samples, accuracy on trigger-transparent samples and difference on accuracy between genuine and manipulated DNN respectively, while columns revealing results with and without autoencoder-powered trigger generation strategy on image and ECG respectively. 
As we can see from the table, in image tests, the success rate drops sharply from $95.4\%$ to $0\%$ on original and $98\%$ to $0\%$ on external without using autoencoder-powered trigger generation mechanism when facing input pre-processing defenses. In ECG tests, the success rate drops from $93.3\%$ to $45.4\%$ on original and $97.5\%$ to $33.8\%$ on external without autoencoder-powered trigger generation mechanism when facing input pre-processing defenses. The decrease of success rate for ECG is less than the image due to the inherently noisy nature of time-series data, which reveals that the noise contained in time series data might contribute to masking the trigger. The accuracy evaluation on both image and ECG shows a slightly increase from while reducing accuracy from $73.3\%$ to $74.1\%$ and $75.8\%$ to $76.3\%$ respectively after applying autoencoder-powered trigger generation mechanism. From the accuracy and difference evaluations, we can see that the implementation of autoencoder-powered trigger generation can also ensure the stealthiness of the manipulated model. The experimental results reveal that the autoencoder-powered trigger generation strategy can defeat pruning-based defenses while maintaining considerable stealthiness.
\begin{table*}[!htb]
\centering
\caption{Comparison with and without autoencoder-powered trigger generation strategy}
\label{my-label}
\begin{tabular}{|l|l|l|l|l|}
\hline
         & image($\mathcal{A+B+C}$) & image($\mathcal{A+C}$) & ECG($\mathcal{A+B+C}$) & ECG($\mathcal{A+C}$) \\ \hline
$SR_O$   & $95.4\%$     & $0\%$    & $93.3\%$   & $45.4\%$ \\ \hline
$SR_E$   & $98\%$      & $0\%$    & $97.5\%$   & $33.8\%$ \\ \hline
Accuracy & $74.1\%$     & $73.3\%$   & $77.6\%$   & $75.8\%$ \\ \hline
$Dif_A$ & $1.7\%$     & $2.5\%$   & $3.4\%$   & $4.2\%$ \\ \hline
\end{tabular}
\end{table*}

To evaluate the effectiveness of the defense-aware retraining algorithm, we compare the performance of our algorithm with ($\mathcal{A+B+C}$) and without ($\mathcal{A+B}$) such strategy, when the pruning-based and fine-tuning defenses are present. Table 7 demonstrates an example for the image (VGG16) and ECG learning model (2D-CNN). 
\begin{table*}[!htb]
\centering
\caption{Comparison with and without defense-aware retraining strategy}
\begin{tabular}{|c|c|c|c|c|}
\hline
\multirow{3}{*}{} & \multirow{3}{*}{image($\mathcal{A+B+C}$)} & \multirow{3}{*}{image($\mathcal{A+B}$)} & \multirow{3}{*}{ECG($\mathcal{A+B+C}$)} & \multirow{3}{*}{ECG($\mathcal{A+B}$)} \\
                  &                                           &                                         &                                       &                                       \\
                  &                                           &                                         &                                       &                                       \\ \hline
$SR_O$            & $95.4\%$                                  & $31.5\%$                                & $93.3\%$                              & $38.2\%$                              \\ \hline
$SR_E$            & $98\%$                                    & $52.1\%$                                & $97.5\%$                              & $58.4\%$                              \\ \hline
Accuracy          & $74.1\%$                                  & $73.2\%$                                & $77.6\%$                              & $74.2\%$                              \\ \hline
$Dif_A$           & $1.7\%$                                   & $2.6\%$                                 & $3.4\%$                               & $5.8\%$                               \\ \hline
\end{tabular}
\end{table*}

Rows 2-5 show the experimental results of success rate for trojaned samples, accuracy on trigger-transparent samples and difference on accuracy between genuine and manipulated DNN respectively, while columns revealing results with and without defense-aware retraining strategy on image and ECG respectively. 
The experimental results demonstrate that success rate decrease from $95.4\%$ to $31.5\%$ on original image datasets and $98\%$ to $52.1\%$ on external image datasets without using defense-aware retraining mechanism on when facing pruning-based defenses while reducing accuracy from $74.1\%$ to $73.2\%$. From the results on ECG, we can see that success rate declines from $93.3\%$ to $38.2\%$ on original and $97.5\%$ to $58.4\%$ on external without using defense-aware retraining mechanism when facing pruning-based defenses, while the accuracy drops from $76.3\%$ to $74.2\%$. The experimental results reveal that the defense-aware retraining strategy can defeat pruning-based defenses while maintaining considerable stealthiness.
\begin{figure}[!htb]
\centering
\includegraphics[width=3.5in,height=1.5in]{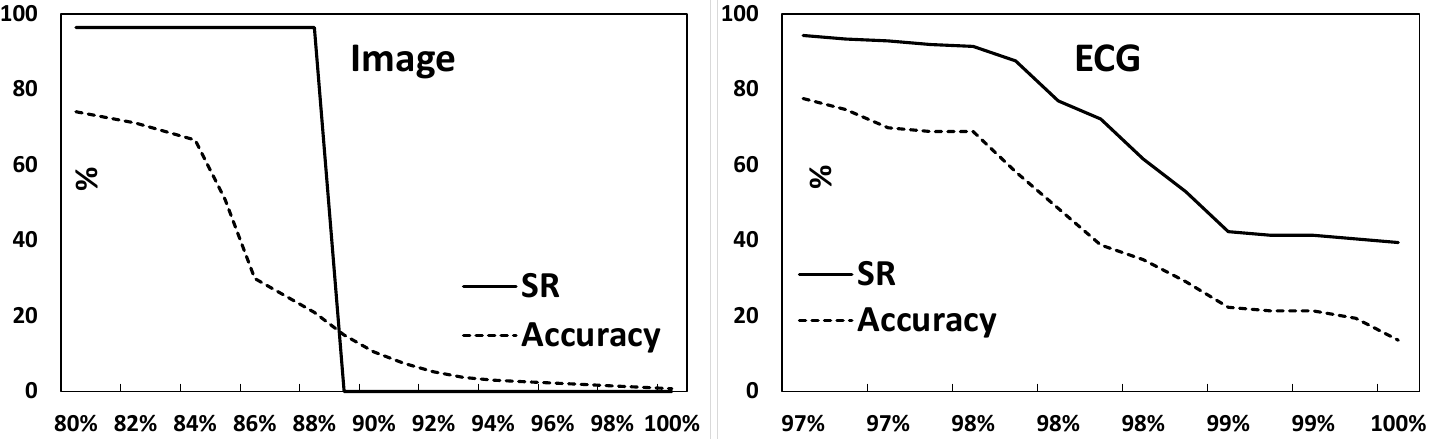}
\caption{Success rate and accuracy evaluation versus proportion of pruned neurons.}
\label{fig:overview}
\end{figure}

Next, we demonstrate how the defensive retraining is able to evade the pruning defense according to an additional dimension, i.e., the proportion of pruned neurons. 
Figure 9 describes the success rate and accuracy evaluation versus the proportion of pruned neurons for our enhanced backdoor attacks on image and ECG recognition when facing pruning-based defenses. 
The success rate and accuracy for both image and ECG show a decline as the fraction of pruned neurons increase. 
For the image-based test, the upsurge of the proportion of pruned neurons causes a sharp drop in the success rate, which reveals the effectiveness of the pruning-based defense. However, the advantage of our enhanced backdoor attack is that the resilience (reflecting by the drop of success rate) against defense-aware attack is at the cost of accuracy. For example, the significant decline (from $95.4\%$ to approximately 0) of success rate happens only when the accuracy on clean inputs decreases below $20\%$ after pruning more than $86\%$ neurons. We also conduct the baseline attack without ranking-based selection and defense-aware retraining, in which the success rate declines to $10\%$ with only $5\%$ reduction in classification accuracy. 

The degrees of pruning on the backdoored ECG causes both the accuracy and the success rate to fall as neurons are pruned gradually. As shown, to reduce the success rate of our attack from $93.3\%$ to $40\%$, the pruning defense has to reduce the accuracy to below $40\%$. 

Furthermore, we also evaluate the difference of performance when conducting full (\textbf{Type I Adversary}) or partial tuning (\textbf{Type II Adversary}), i.e., $f ' +g$ or $ f'+g ' $, in various setting respectively. Figure 10 shows the change of success rate when varying the threshold of $ \eta $ (the adjustment magnitude or learning rate) and the percentage of adjusted parameters $\theta$ respectively. In both cases, we consider the settings that the DNN is perturbed in $f ' +g$ or $ f'+g ' $ manner respectively. 
\begin{figure}[!htb]
\centering
\includegraphics[width=3.5in,height=1.5in]{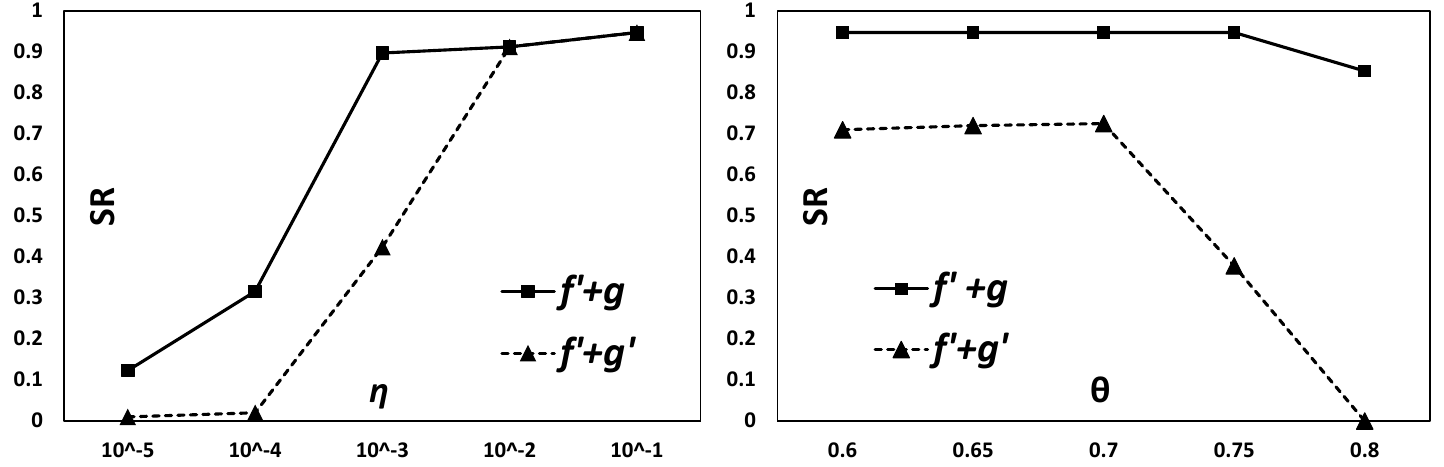}
\caption{Success rate versus the threshold of adjustment amplitude and positive/negative impact.}
\label{fig:overview}
\end{figure}

We make the following three observations. 
(1) As shown as the left subfigure, the success rate of the attack increases as the $ \eta $ becomes larger for both $ f'+g'$ and $ f'+g$ tuning, revealing that a larger adjustment amplitude provides more manipulation space for the adversary. And a relatively small adjustment amplitude can lead to a considerable high success rate, e.g., $\eta =10^{-3}$ enables the adversary to force the system to misclassify more than $80\%$ and $40\%$ of the trojaned inputs using $ f'+g $ and $ f'+g' $ tuning respectively. It means that the imperceptible adjustment magnitude used to conduct our attack is easy to be hidden in the variance of pre-trained DNNs, which can reflect the easiness of our attack. 
(2) As shown in the right subfigure, the success rate of the attack declines as the as $\theta$ increases for both $ f'+g'$ and $ f'+g$ tuning. Compared with the amplitude of the adjustment parameters, the amount of them tends to have less impact on the success rate. This can reveal that the enhanced backdoor attack is effective even under the setting of extremely low perturbation amplitude and full-system tuning.
(3) As demonstrated in both subfigures, the partially tuning, i.e., $f'+g$, can work more efficiently than full-system tuning, i.e., $f'+g'$, whenever varying the amplitude or number of adjustment parameters. For a suitable $\eta$ and $\theta$, the success rate can be achieved at $98\%$, however, the $f'+g'$ tuning might be disabled in some case. 

\subsubsection{ Feasibility and Easiness of Attack}
Table 8 shows the neuron selection time (Column NS), trigger generation time (Column TG) and retraining time (Column RT). 
As shown from the table, it takes less than 20 min to select neuron and 15 minutes to generate triggers for both image and time series data. The time of retraining the model is related to the internal layer and the size of the model. Generating inverse engineering input is the most time consuming step as all possible output results should be considered. Depending on the size of the model, the time varies from one hour to nearly days.
\begin{table}[!htb]
\centering
\caption{Time-based evaluation}
\label{my-label}
\begin{tabular}{|l|l|l|l|}
\hline
Time (min)  & NS & TG & RT \\ \hline
1D-CNN-ECG  & 4.5                & 3.1                 & 26         \\ \hline
2D-CNN-ECG  & 3.3                & 2.8                & 26         \\ \hline
ResNet-Brain  & 5.6                & 3.4               & 48         \\ \hline
VGG16-image & 19.1               & 14.8               & 250        \\ \hline
 \end{tabular}
\end{table}
\section{Related Works}
Poisoning attacks, in which the attack is conducted by poisoning the training data to consequently force the system to generate targeted/non-targeted wrong prediction (alter the behavior of a model) \cite{biggio2012poisoning,xiao2015support}. 
Poisoning attacks on machine learning models have also been studied by many researchers \cite{munoz2017towards,xiao2015feature,yang2017generative}. Xiao et al. \cite{xiao2015feature} demonstrate a poisoning attack on common feature extractor models. A way to poison DNNs with back-gradient optimization is proposed by \cite{munoz2017towards}. Yang et al. \cite{yang2017generative} use the generative method to poison DNN. Poisoning attacks focus on causing the poisoned models to misbehave under normal input while our DNN backdoor attack focuses on making the manipulated DNNs behave normally under normal input and behave as what the attacker desires under input with a trigger. 
Backdoor attacks, in which the attack is performed by adjusting the input data at inference time to trigger the system to maliciously behave \cite{ liu2017trojaning, gu2019badnets}. The difference is that the poisoning attack does not rely on any trigger, and manipulates the behavior of a specific model on a set of clean samples. BadNet is proposed in \cite{gu2019badnets} to inject a backdoor to a deep neural network via manipulating the training dataset. A target label and a trigger pattern are initially designed, followed by subsequent training on a subset of training images with the trigger and targeted labels. Liu et al. proposed a backdoor attack that requires less access to the training data\cite{liu2017trojaning}, by constructing triggers that induce significant responses at some neurons in the DNN model. A more restricted scenario is considered in \cite{chen2017targeted} when conducting a backdoor attack, where the attacker can only modify a limited proportion of the training dataset. A stealthier variant of backdoor attacks, called latent backdoor attack, is proposed in \cite{yao2019latent} by embedding backdoor at latent representation. \cite{shafahi2018poison,yao2019latent} also conducts a back attack that targets the transfer learning scenario. Student training is affected by manipulating poisoned images based on features extracted from the Teacher model. 

Several defense approaches have been proposed to strengthen the aforementioned attacks. Defenses against poisoning attacks mostly focus on sanitizing training data and removing poisoned samples.
To defend perturbation attacks, researchers \cite{meng2017magnet,papernot2015distillation,xu2017feature} propose several defense methods. Papernot et al. \cite{papernot2015distillation} use distillation in the training procedure to defend perturbation attacks. Xu et al. \cite{xu2017feature} defend perturbation attacks through feature squeezing which reduces the bit color or smooth the image using a spatial filter and thus limits the search space for perturbation attack. Meng et al. \cite{meng2017magnet} propose a mechanism to defend the black box and grey box adversarial attacks. Carlini et al. \cite{carlini2016towards} demonstrate how to bypass distillation defense and ten other different defense mechanisms. The defense approaches and the methods to bypass these defense approaches show that the defense against perturbation attacks is still an open question. Another common defense method is adversarial training \cite{ganin2016domain,shrivastava2017learning}. Liu et al. \cite{liu2018fine} proposed Fine-Pruning based defense to remove backdoor triggers by first pruning redundant neurons that are the least useful for classification, then fine-tuning the model using clean training data to restore model performance.

Our proposed attack differs by considering strong defense approaches that exist and investigating the effects of the attack with/without access to the Student model.
To our best knowledge, this work is the first attempt that studies enhanced backdoor attack to defeat pruning based, fine-tuning/ retraining based and input preprocessing defenses, demonstrating on both image and time-series data and comparing both full-system tuning and partial-system tuning.

\section{Conclusion}
With the prevalence of sharing and using public pre-trained models, attackers have many new opportunities, e.g., performing a backdoor attack to manipulate the host system using these pre-trained models.
In this paper, we took an initial step towards conducting an enhanced backdoor attack on both image and time-series data-based learning systems, when facing three strong defenses. 
We first addressed the feasibility of the attack under more realistic constraints while defeating commonly-adopted defenses, i.e., some strong defenses might have been implemented, the generating and perturbation processes should be fast and easy to conduct, and the original training datasets are unavailable due to privacy or copyright concerns. 
Therefore, three optimization strategies are used to generate triggers and retrain DNNs, e.g., \textit{Ranking-based Neuron Selection, Autoencoder-powered Trigger Generation and Defense-aware Retraining}. 
We conducted the evaluation and case studies on real-world images, MRI image and ECG applications to show that the attack is effective against pruning based, fine-tuning/retraining based and input pre-processing based defenses, as well as being feasible and easy for the adversary to launch such attacks. The experiments demonstrated that our enhanced attack can maintain the same classification accuracy as a genuine model on clean input while ensuring a high attack success rate on trojaned input incorporated with our designed trigger. The experiments reveal that our enhanced attack can maintain the high classification accuracy as a genuine model on clean inputs while improving attack success rate on trojaned inputs in the presence of pruning-based and/or retraining-based defenses. 

A few possible future extensions include: First, we enhance the detection evasiveness of our attack approach so that the crafted model can be more indistinguishable from the genuine one. Second, implementing and evaluating a more strong and feasible defense is an interesting future work. Finally, besides the backdoor attacks, we will consider other attacks and threats (e.g., adversarial example attack or privacy concerns).
\bibliographystyle{IEEEtran}
\bibliography{bare_jrnl_compsoc}
\renewcommand{\baselinestretch}{0.8}
%



%




\begin{IEEEbiography}[{\includegraphics[width=1in,height=1.25in,clip,keepaspectratio]{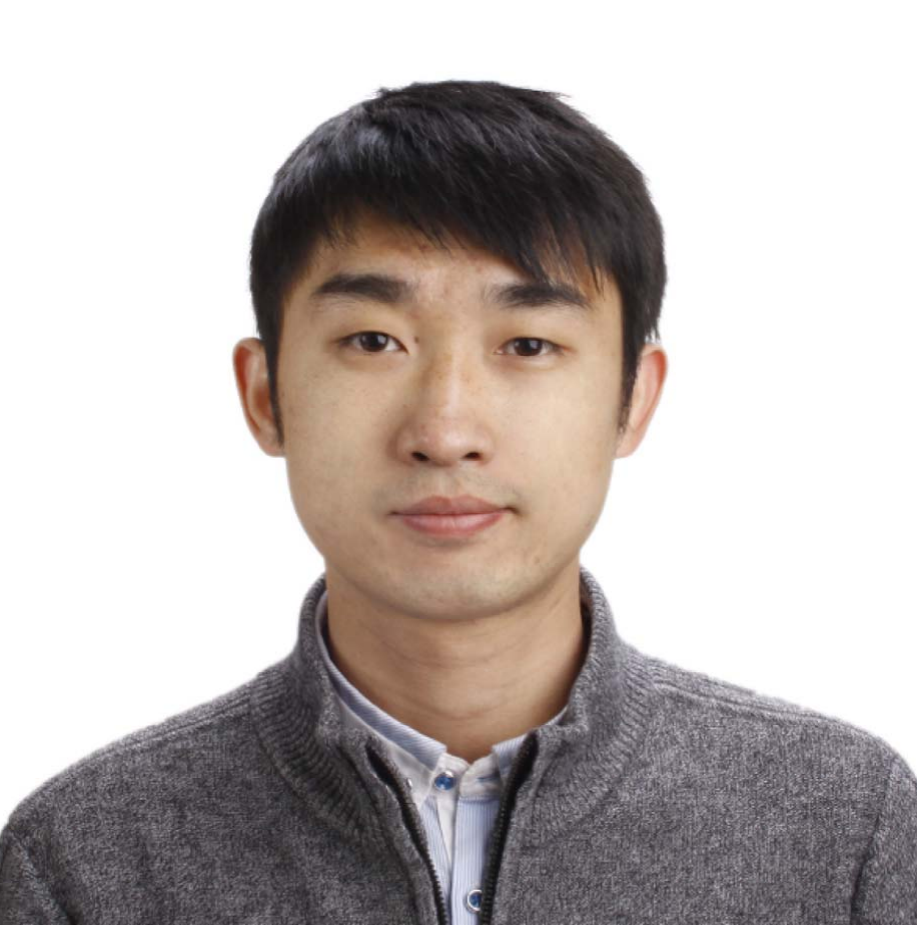}}]{Shuo Wang}
Dr. Shuo Wang is a Joint Research Fellow in the Faculty of Information Technology at Monash University and CSIRO's Data61. Before joining Monash, his Ph.D. and earlier research work was in the School of Computing and Information Systems, University of Melbourne. His main research interests include the areas of: 
Adversarial Machine Learning: attacks and defenses of deep neural networks;
Computer security and privacy, including security and privacy issues in systems, networking, and databases.
\end{IEEEbiography}

\begin{IEEEbiography}[{\includegraphics[width=1in,height=1.25in,clip,keepaspectratio]{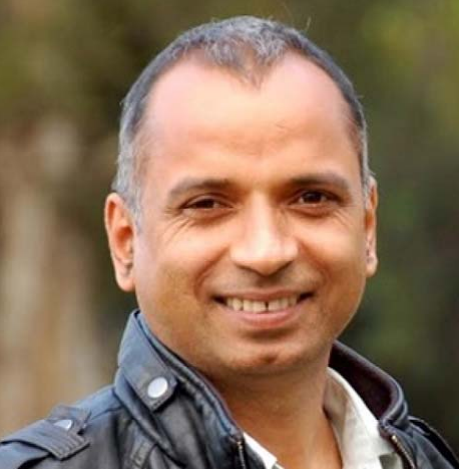}}]{Surya Nepal}
Dr. Surya Nepal is a Principal Research Scientist working on trust and security aspects of Web Services at CSIRO's Data61. His main research interest is in the development and implementation of technologies in the area of Web Services and Service-Oriented Architectures. At CSIRO, Surya researched in the area of multimedia databases, Web Services and Service-Oriented Architectures, security, privacy and trust in collaborative environment. 
\end{IEEEbiography}

\begin{IEEEbiography}[{\includegraphics[width=1in,height=1.25in,clip,keepaspectratio]{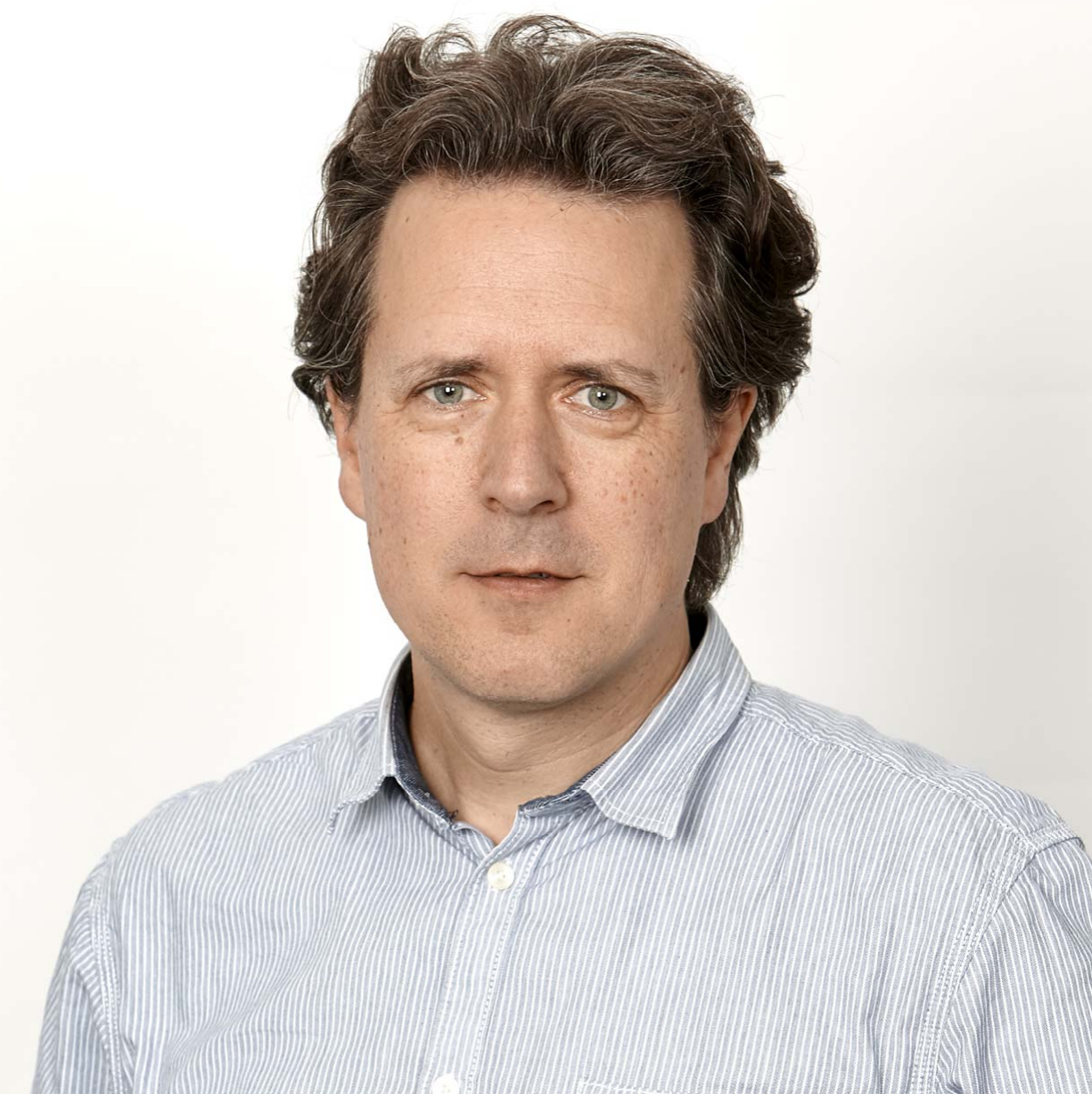}}]{Carsten Rudolph}
Carsten Rudolph is an Associate Professor of the Faculty of IT at Monash University, Melbourne, Australia, and Director of the Oceania Cyber Security Centre OCSC. Carsten Rudolph has contributed extensively to four key areas of cybersecurity: Trusted Computing, Security of critical infrastructures and Security of IT Networks; Security by design/security engineering / formal methods for security; Validation and design of security protocols; Digital forensic readiness and secure digital evidence.
\end{IEEEbiography}
\begin{IEEEbiography}[{\includegraphics[width=1in,height=1.25in,clip,keepaspectratio]{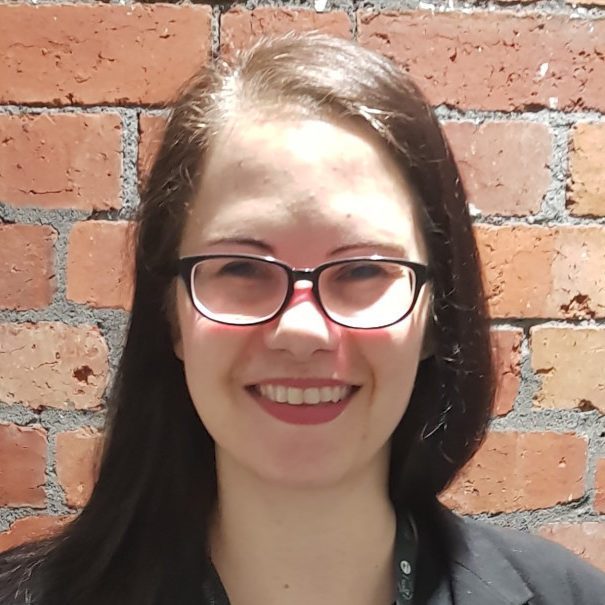}}]{Marthie Grobler}
Dr. Marthie Grobler currently holds a position as Senior Research Scientist at CSIRO's Data61 in Melbourne, Australia where she leads the human-centric cybersecurity team, driving the work on cybersecurity governance, policies, and awareness.
\end{IEEEbiography}
\begin{IEEEbiography}[{\includegraphics[width=1in,height=1in,clip,keepaspectratio]{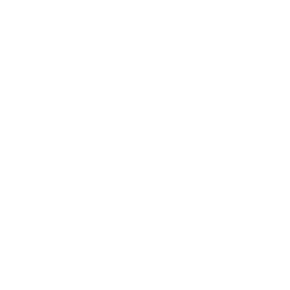}}]{Shangyu Chen}
Shangyu Chen is a postgraduate student in the Faculty of Computing and Information Systems at the University of Melbourne, Melbourne, Australia. His research topic is machine learning and security.
\end{IEEEbiography}
\begin{IEEEbiography}[{\includegraphics[width=1in,height=1in,clip,keepaspectratio]{kong}}]{Tianle Chen}
Tianle Chen is a postgraduate student in the Faculty of Information Technology at Monash University, Melbourne, Australia. His research topic is deep learning and privacy.
\end{IEEEbiography}
\end{document}